\documentclass[letterpaper]{article}
\usepackage[preprint]{aaai2027}
\usepackage[hyphens]{url}
\usepackage{graphicx}
\usepackage{natbib}
\usepackage{caption}
\usepackage{amsmath,amssymb}
\usepackage{algorithm}
\usepackage{algorithmic}
\usepackage{booktabs}
\usepackage{tabularx}
\usepackage{array}

\newcolumntype{Y}{>{\raggedright\arraybackslash}X}

\title{STRATA: Self-Learning Through Role-Aligned Tiered Agents for Real-Time Strategy Games}
\author{
Xinhe Tian\textsuperscript{1}, Xiaoyue Zhang\textsuperscript{1},
Ziyou Zhang\textsuperscript{2}, Jiacheng Li\textsuperscript{3},\\
Xiaoqiang Jin\textsuperscript{2}, Qianchuan Zhao\textsuperscript{4},
Gaochen Cui\textsuperscript{5,*}
}
\affiliations{
\textsuperscript{1}School of Instrumentation Science and Optoelectronic Engineering, Beijing University of Aeronautics and Astronautics\\
\textsuperscript{2}Tsinghua University; \textsuperscript{3}University of Chinese Academy of Sciences\\
\textsuperscript{4}Department of Automation, Tsinghua University; \textsuperscript{5}CFINS, Tsinghua University\\
\textsuperscript{*}Corresponding author
}

\begin{document}
\maketitle

\begin{abstract}
Real-time strategy (RTS) games require agents to coordinate economic development, production and construction, base defense, unit organization, and attack timing over long matches. Existing studies have applied large language models to command decision-making in RTS games, enabling agents to read textual game states and generate high-level plans. However, long inference latency can cause them to miss critical tactical events. The complexity and tactical diversity of full RTS matches also leave existing systems heavily dependent on manually written experience-based prompts, with limited ability to learn continuously from past games. We present STRATA, a role-aligned hierarchical system with cross-game self-learning for Red Alert. STRATA assigns in-game strategic, logistical, and tactical decisions to a Strategic Agent (SA), Logistics Agent (LA), and Tactical Agent (TA), respectively. The SA generates high-level directives based on the global game state and relevant experience cards, while the LA and TA handle logistics and tactical execution. After each match, a Review Agent (RA) derives candidate experience from game traces, validates and revises it using evidence from subsequent matches, and compresses strategic experience supported across multiple games into concise experience cards for SA retrieval. We evaluate STRATA through the formation of experience cards, full-match comparisons before and after learning, and experience learning against AI opponents with different play styles. Under a fixed scenario, using the learned experience cards increases the observed win rate from 30\% to 100\%. Sequential learning against AI opponents with different play styles also produces distinct long-term strategic experience.
\end{abstract}

\section{Introduction}

Real-time strategy (RTS) games require agents to coordinate economic development, production, technology, base defense, and unit operations under partial observability \citep{ontanon2013survey,vinyals2017sc2le,samvelyan2019smac}. Resource investments determine future production capacity. Defensive pressure can disrupt expansion and technology schedules, while attack timing depends on current forces, reinforcements, and the opponent's state. The trade-off among economy, technology, and military production is therefore central to RTS strategy \citep{ontanon2013survey,robertson2014review}. Many decisions reveal their effects only minutes later, and opponents continually change the conditions under which earlier plans were made. Agents must therefore revise their strategies over long horizons \citep{vinyals2017sc2le,vinyals2019grandmaster}. RTS games combine partial observability, large state and action spaces, long decision chains, and delayed feedback \citep{vinyals2017sc2le,samvelyan2019smac,vinyals2019grandmaster}. Their explicit rules, traceable processes, and measurable outcomes provide a controlled setting for evaluating long-horizon planning, online decision-making, and real-time execution \citep{vinyals2017sc2le,samvelyan2019smac,andersen2018deeprts}. StarCraft II has been widely used for intelligent-agent research \citep{vinyals2017sc2le}, while Red Alert has recently been introduced as a diagnostic benchmark for long-horizon LLM decision-making \citep{li2026winning}.

Reinforcement learning has long been a major approach for building game-playing agents. Early deep reinforcement learning systems learned control policies directly from pixels \citep{mnih2015human}. Later systems combined search, self-play, and large-scale distributed training for Go, chess, shogi, Dota 2, and StarCraft II \citep{silver2016mastering,silver2018general,berner2019dota,vinyals2019grandmaster}. However, these systems often require extensive environment interaction, sustained self-play, and substantial computation \citep{silver2018general,berner2019dota,vinyals2019grandmaster}. Large language models provide another approach. Gato addressed tasks across games, text, and robotic control \citep{reed2022gato}, while CICERO combined language communication with strategic reasoning in a long-horizon game that involves both cooperation and competition \citep{bakhtin2022cicero}.

\begin{figure*}[t]
\centering
\includegraphics[width=0.82\textwidth]{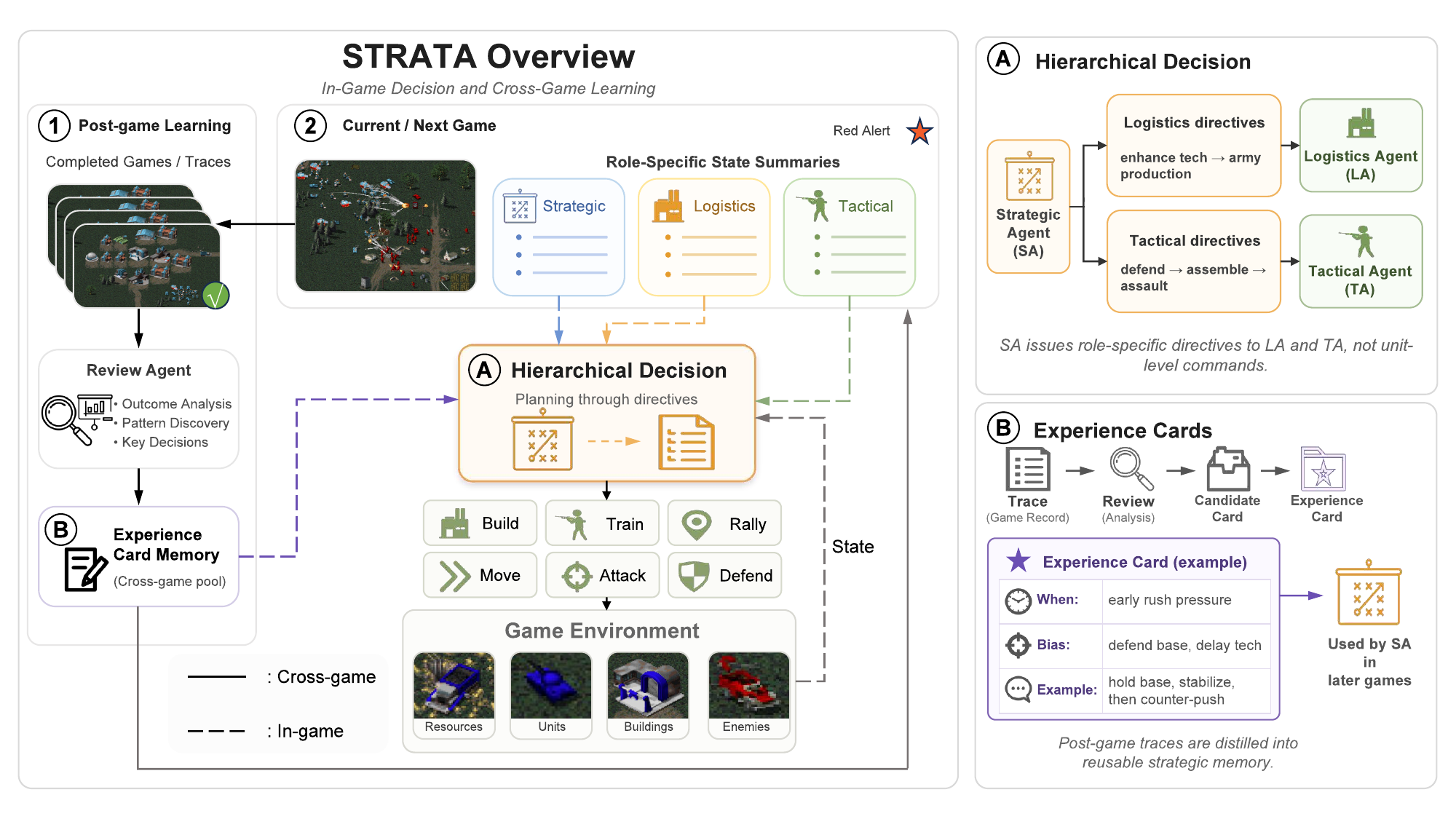}
\caption{Overview of STRATA. The system combines role-aligned hierarchical decision-making during each match with a closed loop for post-game experience formation, cross-game validation, and retrieval in later matches.}
\label{fig:strata-overview}
\end{figure*}

Recent studies use large language models for command decision-making in RTS games. Textual observations, state summaries, and action interfaces allow these models to make high-level economic, technological, and combat decisions \citep{ma2023textstarcraft,li2024llmpysc2,shao2024swarmbrain}. However, resources, production queues, unit positions, and local engagements may all change during a single inference cycle. A model that handles strategy, logistics, and combat must also process a broader input and reasoning scope. TextStarCraft II, LLM-PySC2, and SwarmBrain improve sequential state summaries, interaction interfaces, and rapid tactical responses, respectively \citep{ma2023textstarcraft,li2024llmpysc2,shao2024swarmbrain}. Even so, model decision cycles can still lag behind real-time game changes.

Existing systems also depend heavily on manually written experience-based prompts. TextStarCraft II identifies the lack of feedback loops and long-term memory as important limitations \citep{ma2023textstarcraft}. Reflection, language memory, and skill reuse can exploit past feedback, but these methods have mainly been evaluated in open-ended environments such as Minecraft. Compared with full RTS matches, such environments involve less continuous adversarial interference. Their stored experience therefore needs less frequent revision as opponent strategies and game phases change.

To address these problems, we present STRATA (Self-Learning Through Role-Aligned Tiered Agents for Real-Time Strategy Games), a role-aligned hierarchical system with cross-game self-learning for full RTS matches. STRATA includes three in-game agents and one post-game review agent. The Strategic Agent (SA) makes strategic decisions and sends high-level directives to the Logistics Agent (LA) and Tactical Agent (TA). The LA and TA use role-specific observations to execute logistics and tactical actions, respectively. The three in-game agents differ in information scope, decision authority, and update frequency. This separation allows strategic, logistical, and tactical decisions to run independently. It also prevents low-frequency strategic inference from serially blocking logistics and tactical execution. After each match, the Review Agent (RA) examines the game trace and current experience state. It proposes new candidate experience or revises existing experience. Strategic guidance supported across multiple matches is then compressed into concise experience cards for the SA to retrieve in later matches.

We evaluate STRATA in Red Alert through experience-card formation, full-match comparisons before and after learning, and experience learning against Rush AI and Turtle AI. Across the experiments, the RA continually forms and revises strategic experience from match traces, producing a formal experience library with 16 cards. When the SA uses these cards, the observed win rate increases from 30\% to 100\%. The system also develops distinct long-term experience for AI opponents with different play styles.

In summary, our main contributions are as follows:
\begin{enumerate}
\item We propose a role-aligned hierarchical architecture composed of an SA, LA, and TA. The architecture separates strategic, logistical, and tactical decisions by information scope, decision authority, and update frequency.

\item We propose an RA-driven cross-game experience-learning method for the SA. The RA uses evidence from subsequent matches to validate and revise candidate experience continuously, then accumulates supported strategic experience in experience cards.

\item We evaluate STRATA from three perspectives: experience-card formation, full-match performance before and after learning under a fixed setting, and sequential experience learning against Rush and Turtle opponents. These experiments show how experience cards are formed, how they affect full-match performance, and how they evolve under different opponent styles.
\end{enumerate}

\section{Related Work}

\textbf{LLM agents in interactive environments.}
Large language model agents extend language-based reasoning to continuous interaction with an environment. Chain-of-Thought exposes intermediate reasoning steps, ReAct alternates reasoning and action, and Tree of Thoughts and LATS search over alternative reasoning and action paths \citep{wei2022cot,yao2023tot,yao2023react,zhou2023lats}. These methods support goal decomposition and multi-step planning. However, they are mainly evaluated in textual, web-based, or well-structured tasks in which the environment usually waits for inference to finish.

Research on games also studies long-horizon planning in interactive environments. MineDojo supports open-ended Minecraft tasks, while GITM decomposes long-term goals into subtasks and actions \citep{fan2022minedojo,zhu2023gitm}. ProAgent combines independent planning with teammate-intention inference for decentralized cooperation \citep{zhang2024proagent}. Other studies organize multiple language-model agents through hierarchies and specialized roles \citep{ahn2025hima,deng2024smacr1,qi2025masmp}. These works demonstrate task allocation and information exchange, but they mainly study open-ended exploration, cooperation, or relatively stable interactions.

\textbf{LLM agents for RTS games.}
TextStarCraft II organizes economic, technological, and combat states through single-frame and multi-frame summaries for sequential high-level planning \citep{ma2023textstarcraft}. LLM-PySC2 expands action spaces, multimodal observations, and multi-agent interfaces for complete StarCraft II tasks \citep{li2024llmpysc2}. These systems make LLMs easier to integrate with RTS environments. However, state representations still trade information coverage for input length: concise summaries may omit local changes, while richer observations and action spaces increase the reasoning burden. A single model may therefore struggle to operate across different decision time scales.

SwarmBrain adds rapid-response mechanisms to high-level planning so that some local events do not need to wait for the next full inference \citep{shao2024swarmbrain}. However, its triggers and behaviors require manual design and remain limited by rule coverage. Other work adds domain knowledge through adaptation and expert prompts \citep{khan2024scphi2,li2025hep}, preserves state through visual or structured observations \citep{ma2026ava}, and improves action executability through hierarchies, role-based decisions, code policies, and state machines \citep{ahn2025hima,deng2024smacr1,qi2025masmp}. These approaches improve state understanding, action generation, or response speed. However, major decisions often remain in one core agent or control loop and still require manual maintenance, larger contexts, or predefined rules. Strategy, logistics, and tactics differ in both information needs and rates of change. How to separate their time scales without blocking local execution remains underexplored.

\textbf{Reflection, memory, and experience learning.}
Natural-language feedback and external memory allow large language models to reuse experience without parameter updates. STaR constructs examples from successful reasoning, Self-Refine repeatedly revises outputs using model feedback \citep{zelikman2022star,madaan2023selfrefine}, and Reflexion stores verbal reflections after failures for later attempts \citep{shinn2023reflexion}. These methods can reduce repeated errors, but their reflections often rely on a limited number of trajectories. Incorrect attribution or changing conditions may therefore cause the agent to rely too heavily on one explanation.

ExpeL extracts reusable experience by comparing successful and failed trajectories \citep{zhao2024expel}, while Generative Agents combine retrieval, reflection, and planning over time \citep{park2023generative}. GITM, Voyager, and JARVIS-1 use task histories, skill libraries, and multimodal memory for long-term exploration in Minecraft \citep{zhu2023gitm,wang2023voyager,wang2023jarvis}. These studies show that past experience can guide later decisions, but they usually assume relatively stable goals. In full RTS matches, strategic effects vary with opponent behavior, economy, game phase, and force composition. Conclusions from a single match therefore require continued correction through later supporting evidence and counterexamples.

\section{STRATA}

\subsection{System Overview}

In a full RTS match, strategic judgment, logistics planning, and local combat have different information needs and rates of change. STRATA therefore organizes in-game decision-making as a role-aligned hierarchical architecture composed of a Strategic Agent (SA), Logistics Agent (LA), and Tactical Agent (TA). The SA uses the global game state and relevant experience to set priorities among economy, technology, force accumulation, and attack timing. It then sends high-level directives to the LA and TA. The LA handles construction, production, repair, and resource allocation, while the TA handles defense, rallying, advances, and local combat. Each agent uses role-specific observations and operates independently at its own time scale. Logistics and tactical execution therefore do not need to wait for every low-frequency strategic inference to finish. Figure~\ref{fig:strata-overview} summarizes the in-game hierarchy and the cross-game experience-learning loop.

After each match, a Review Agent (RA) reads the complete game trace and current experience state. It organizes strategic evidence and proposes new candidate experience or revisions to existing experience. Experience supported by later matches is compressed into concise experience cards for the SA to retrieve in subsequent matches. The RA updates cross-game experience, the SA uses this experience during a match, and the LA and TA execute logistics and tactical actions. In this way, experience cards transfer findings from completed matches into later strategic decisions.

\subsection{Role-Aligned Hierarchical In-Game Decision-Making}

A full RTS state contains economic, technological, production, force-level, and local-combat information. STRATA derives three role-specific observations from the shared game state according to the needs of each decision type. Each in-game agent can act only within its assigned decision authority. Let $s_t$ denote the shared game state at time $t$, and let $U_t^{\mathrm{TA}}$ denote the set of combat units controlled by the TA. The observations are

\begin{equation}
\begin{aligned}
o_t^{\mathrm{SA}}&=\phi_{\mathrm{SA}}(s_t),\qquad
o_t^{\mathrm{LA}}=\phi_{\mathrm{LA}}(s_t),\\
o_t^{\mathrm{TA}}&=\phi_{\mathrm{TA}}(s_t,U_t^{\mathrm{TA}}).
\end{aligned}
\end{equation}

The SA observes the economic phase, power state, technology progress, total force strength, opponent pressure, relevant experience cards, and compressed logistics and tactical states. The LA observes cash, construction prerequisites, production queues, available production options, repair needs, and current production capacity. The TA receives only its assigned combat units, local opponent information, base positions, and current tactical objective. Each observation contains the information needed for strategic planning, logistics decisions, or local combat. This design keeps each agent's reasoning within its assigned role.

Role boundaries also constrain agent outputs. The SA issues high-level directives to the LA and TA. The LA manages construction, production, repair, and rally-point settings, while the TA controls its assigned combat units. Their decision process is

\begin{equation}
\begin{aligned}
(d_t^{\mathrm{LA}},d_t^{\mathrm{TA}})
&=\pi_{\mathrm{SA}}(o_t^{\mathrm{SA}},C_t),\\
a_t^{\mathrm{LA}}
&=\pi_{\mathrm{LA}}(o_t^{\mathrm{LA}},d_t^{\mathrm{LA}}),\\
a_t^{\mathrm{TA}}
&=\pi_{\mathrm{TA}}(o_t^{\mathrm{TA}},d_t^{\mathrm{TA}}),
\end{aligned}
\end{equation}

where $C_t$ is the set of experience cards relevant to the current situation, and $d_t^{\mathrm{LA}}$ and $d_t^{\mathrm{TA}}$ are the high-level logistics and tactical directives. The SA selects the current strategic priority among economy, technology, force accumulation, and attack timing. The LA and TA then choose executable actions from their latest observations.

For example, suppose the SA decides to establish a technology base while withstanding early pressure. It can direct the LA to maintain power and basic production capacity, and direct the TA to protect the base and ore fields. The LA chooses a concrete construction and production order based on current cash, queues, and building conditions. The TA decides whether to defend, rally, or counterattack based on opponent positions and available units. Both agents adapt their execution around the same strategic objective while retaining decision space within their roles. Role-specific observations also reduce the input scope of each model call. The SA, LA, and TA can therefore focus on global strategy, logistics, and the local battlefield, respectively.

The SA, LA, and TA operate asynchronously at different time scales. The Red Alert environment continues to advance and publish the latest state. Each agent maintains its current valid output and tracks any pending model request. The SA reassesses the global strategy at a lower frequency. The LA periodically reads the latest economic and production state and updates its plan when it receives a new logistics directive. The TA replans when it receives a new tactical directive, while its current plan remains active until the update is complete.

For each in-game agent $r\in\{\mathrm{SA},\mathrm{LA},\mathrm{TA}\}$, the decision trigger is summarized as

\begin{equation}
\operatorname{decide}_r(t)=
\begin{cases}
1, & t-\tau_r^{\mathrm{last}}\geq\Delta_r,\\
1, & r\in\{\mathrm{LA},\mathrm{TA}\}
     \land \operatorname{new}(d_t^r),\\
0, & \text{otherwise},
\end{cases}
\end{equation}

where $\tau_r^{\mathrm{last}}$ is the time when agent $r$ last initiated a decision, and $\Delta_r$ is its regular decision interval. The SA updates its strategic judgment mainly at regular intervals. The LA and TA can also start a new decision when they receive a new role-specific directive.

Algorithm~\ref{alg:runtime} summarizes the asynchronous in-game decision process.

\begin{algorithm}[t]
\caption{STRATA In-Game Hierarchical Decision}
\label{alg:runtime}
\textbf{Input}: Red Alert environment $\mathcal{E}$ and formal experience library $M$\\
\textbf{Output}: Full-match trajectory $T$
\begin{algorithmic}[1]
\STATE Initialize shared state, directives
$(d^{\mathrm{LA}},d^{\mathrm{TA}})$,
and current actions
$(a^{\mathrm{LA}},a^{\mathrm{TA}})$

\WHILE{$\mathcal{E}$ is not terminated}
    \STATE Advance $\mathcal{E}$ using current
    $(a^{\mathrm{LA}},a^{\mathrm{TA}})$
    and publish the latest state $s_t$

    \IF{an SA result is available}
        \STATE Update high-level directives
        $(d^{\mathrm{LA}},d^{\mathrm{TA}})$
    \ENDIF

    \IF{an LA result is available}
        \STATE Update the current logistics action
        $a^{\mathrm{LA}}$
    \ENDIF

    \IF{a TA result is available}
        \STATE Update the current tactical action
        $a^{\mathrm{TA}}$
    \ENDIF

    \IF{SA is triggered and no SA request is pending}
        \STATE Construct $o_t^{\mathrm{SA}}$
        and retrieve relevant experience cards $C_t$
        \STATE LaunchAsync
        $\pi_{\mathrm{SA}}(o_t^{\mathrm{SA}},C_t)$
    \ENDIF

    \IF{LA is triggered and no LA request is pending}
        \STATE Construct $o_t^{\mathrm{LA}}$
        \STATE LaunchAsync
        $\pi_{\mathrm{LA}}
        (o_t^{\mathrm{LA}},d^{\mathrm{LA}})$
    \ENDIF

    \IF{TA is triggered and no TA request is pending}
        \STATE Construct $o_t^{\mathrm{TA}}$
        \STATE LaunchAsync
        $\pi_{\mathrm{TA}}
        (o_t^{\mathrm{TA}},d^{\mathrm{TA}})$
    \ENDIF

    \STATE Append states, directives, actions,
    and outcomes to $T$
\ENDWHILE

\STATE \textbf{return} $T$
\end{algorithmic}
\end{algorithm}

While one agent waits for a model response, the environment and the other agents continue to run with their current valid outputs. Until a new SA result arrives, the previous high-level directives remain active. The LA continues its logistics plan using the latest economic and production state, while the TA controls units with its current tactical plan. This asynchronous mechanism allows low-frequency SA reasoning to proceed in parallel with logistics and tactical execution. Production, construction, and local combat continue to use recent game states while model requests complete independently.

\subsection{Review-Agent-Driven Cross-Game Experience Learning}

STRATA represents cross-game experience as strategic preferences for the SA and stores these preferences as experience cards in a formal experience library. Each card describes several strategic directions and their relative strengths for a specific situation. An experience card is represented as

\begin{equation}
\begin{aligned}
c_i&=\left\langle s_i,\mathcal{B}_i,e_i,w_i,q_i\right\rangle,\\
\mathcal{B}_i&=\left\{(b_{i1},\alpha_{i1}),\ldots,
(b_{im},\alpha_{im})\right\}.
\end{aligned}
\end{equation}

Here, $s_i$ describes the situation in which the experience applies, $b_{ij}$ denotes a possible strategic direction, and $\alpha_{ij}$ gives its preference strength. The field $e_i$ provides a concise example of a high-level directive. The values $w_i$ and $q_i$ denote the card weight and confidence, respectively.

For example, when the agent already has a war factory, an experience card may recommend maintaining basic vehicle production, reserving resources for technology, and keeping a safe power margin. The SA evaluates these preferences together with the current game state. Under high opponent pressure and insufficient forces, it gives higher priority to basic production and cash preservation. When defensive pressure is low and the economy is stable, it can place more weight on technology development and advanced-unit production. These preferences help the SA choose the strategic priority of the current phase from the current pressure, economy, and force conditions.

During a match, the system selects a small set of relevant experience cards from the formal library $M$ according to applicability, weight, and confidence:

\begin{equation}
C_t=
\operatorname{TopK}_{c_i\in M}
\left[
\operatorname{Match}(s_i,\hat{o}_t^{\mathrm{SA}})
\cdot w_i\cdot q_i
\right],
\end{equation}

where $\hat{o}_t^{\mathrm{SA}}$ is the global state summary used for experience matching. The RA maintains the cross-game experience that enters the SA's in-game decision context. The SA combines the current game state with the strategic preferences in $C_t$ and then issues high-level directives to the LA and TA.

After match $e$, the RA organizes strategic evidence $E_e$ from the full trajectory $T_e$. This evidence includes economic and production schedules, opponent pressure, key engagements, changes in high-level directives, experience-card retrieval, and subsequent outcomes. The RA also reads the formal experience library $M_{e-1}$ and candidate pool $P_{e-1}$ saved after the previous match:

\begin{equation}
R_e=
\pi_{\mathrm{RA}}
\left(
E_e,
M_{e-1},
P_{e-1}
\right),
\end{equation}

where $R_e$ is the review result produced by the RA. The RA assigns each finding to a new strategic theme, an existing candidate, or a formal experience card. For an existing theme, it records supporting evidence, counterexamples, or revision evidence. A new theme enters the candidate pool and may be merged with semantically related candidates. Existing themes continue to accumulate evidence across matches. With sufficient support, candidate experience can be promoted to formal experience. Later outcomes can also revise the applicability conditions, strategic preferences, weight, and confidence of formal experience. Repeated high-pressure matches gradually strengthen preferences for basic forces, defense, and cash preservation. Repeated low-pressure matches strengthen preferences for economic expansion, technology, and proactive attacks. Side effects observed in later matches trigger further revisions, allowing the experience library to adapt as game conditions change.

In the next match, the SA retrieves the updated formal experience. It selects relevant cards from the current global state and uses them to issue high-level directives to the LA and TA. After the match, the new trajectory returns to the RA and provides further evidence for candidate and formal experience. STRATA thus turns strategic findings from sequential matches into reusable cross-game experience.

Algorithm~\ref{alg:learning} summarizes the post-game experience update process.

\begin{algorithm}[t]
\caption{Review-Agent-Driven Cross-Game Experience Learning}
\label{alg:learning}
\textbf{Input}: Completed trajectory $T_e$, formal experience library $M$, and candidate pool $P$\\
\textbf{Output}: Updated $M$ and $P$
\begin{algorithmic}[1]
\STATE Extract strategic evidence $E_e$ from $T_e$
\STATE $R_e \gets \pi_{\mathrm{RA}}(E_e,M,P)$

\FORALL{finding $r\in R_e$}
    \IF{$r$ introduces a new strategic theme}
        \STATE Add $r$ to $P$, or merge it with a related candidate
    \ELSE
        \STATE Accumulate supporting, counterexample,
        or revision evidence for the related experience
    \ENDIF
\ENDFOR

\FORALL{candidate $p\in P$}
    \IF{$p$ has sufficient cross-game support}
        \STATE Promote $p$ into $M$
    \ENDIF
\ENDFOR

\FORALL{formal experience $c\in M$}
    \IF{sufficient revision evidence is available}
        \STATE Revise the situation, strategic preferences,
        weight, or confidence of $c$
    \ENDIF
\ENDFOR

\STATE \textbf{return} $M,P$
\end{algorithmic}
\end{algorithm}

\section{Experiments}

\subsection{Experimental Setup}

All experiments are conducted in an agent environment built
on the OpenRA engine \citep{openra2026} and its Red Alert mod.
The environment provides structured game-state observations
and executable action interfaces. We use GLM-5.1 as the decision model in every experiment and keep all other environment and runtime settings unchanged. The supplementary material reports the map, scenario seed, agent update intervals, and software versions. We evaluate three AI opponents. Normal AI follows a relatively balanced development and combat style. Rush AI emphasizes rapid early attacks, whereas Turtle AI prioritizes base defense, economic growth, and late-game technology.

We first run 80 sequential matches between STRATA and Normal AI. After each match, the RA updates the candidate and formal experience. We then run ten full matches before learning and ten after learning. Before learning, the SA receives no experience cards. After learning, it receives the 16 formal experience cards formed during the 80 matches. The experience state remains fixed during evaluation, and the 80 experience-learning matches are separate from all evaluation matches.

Finally, the Rush and Turtle experiments start from the same 16 formal experience cards and use separate memory spaces. Each condition runs for ten sequential matches, and the RA updates experience after every match. These experiments compare the long-term strategic preferences that emerge under different opponent play styles.

\subsection{Experience-Card Formation}

We first examine whether STRATA can form reusable strategic experience across sequential full matches. The system plays 80 matches against Normal AI and conducts a post-game review after each match using the process trace, final outcome, and current experience state. Early matches reveal several recurring problems: delayed opening production, imbalanced harvester counts, insufficient vehicle production capacity, premature technology development, and poorly timed defensive counterattacks. The system initially forms 10 formal experience cards from these recurring problems. The cards cover the main strategic components of economy, production, technology, defense, and offense. Table~\ref{tab:experience-card} presents an abridged example of the experience-card format.

\begin{center}
\begin{minipage}{\columnwidth}
\centering
{\small
\setlength{\tabcolsep}{3pt}

\begin{tabularx}{\columnwidth}{
    @{}
    >{\bfseries\raggedright\arraybackslash}p{0.17\columnwidth}
    >{\raggedright\arraybackslash}X
    @{}
}
\toprule

id &
barracks\_seed\_basic\_\newline
defense\_001
\\

when &
Infantry production is available and fewer than six infantry
units are currently fielded.
\\

bias &
Train Rifle Infantry: 0.35;\newline
train Rocket Soldiers: 0.40;\newline
preserve the cash floor: 0.60;\;
pause below 300 cash: 0.40;\; $\ldots$
\\

example &
Direct the LA to form a small mixed infantry defense group,
such as three Rifle Infantry and three Rocket Soldiers.
Keep at least 500 cash after queueing, stop below 300 cash,
and avoid continuously filled queues or additional infantry
production buildings.
\\

weight &
0.9
\\

confidence &
1.0
\\

\bottomrule
\end{tabularx}
}
\captionof{table}{An abridged view of an experience card. The complete card representation is provided in the supplementary material.}
\label{tab:experience-card}
\end{minipage}
\end{center}

As more matches are played, early experience is tested in more complex situations. Some rules prove too broad, such as always switching to an attack after completing technology. Other rules identify a useful direction but omit conditions related to the economy, opponent pressure, or current forces. The RA removes or replaces overly general experience and records new problems as candidate cards. Seven candidates accumulate cross-game evidence in later matches and enter the formal experience library. These candidates mainly concern technology transitions, defensive counterattacks, economic expansion, and attack windows. Experience also shifts from isolated action suggestions to longer-term strategic relationships. Examples include limiting early defensive-unit batches to preserve economic resources, advancing technology only after the basic economy and minimum defense become stable, regrouping after defending the base, and expanding ore fields and harvester counts during safe windows.

By the end of the 80-match sequence, the formal experience library contains 16 cards. As the sequence progresses, fewer independent themes require new cards. Updates instead focus on refining the applicability conditions, strategic preferences, and example boundaries of existing cards. These matches reveal more specific side effects, including continuous cash consumption from basic infantry production, technology development that still begins too early, excessive waiting after forces are assembled, and premature attacks with insufficient units. The system revises card applicability, strategic preferences, and example boundaries accordingly. It completes 14 revisions across 9 formal cards. These revisions include limiting production batches, adding cash and queue conditions, delaying technology under low power, and resetting rally and attack conditions.

After 80 matches against Normal AI, STRATA forms 16 formal experience cards covering economy, production, technology, defense, and offense. Later updates mainly refine their applicability conditions and strategic preferences.

\subsection{Full-Match Performance Before and After Learning}

Table~\ref{tab:before_after_results} compares full-match results against Normal AI before and after experience learning. STRATA wins 3 of 10 matches before learning and all 10 matches after learning, increasing the observed win rate from 30\% to 100\%. The mean and median end times decrease from 28,173.1 and 22,800.0 ticks to 20,842.8 and 17,380.5 ticks, respectively. Both conditions use the same environment configuration and fixed experience state; the only difference is whether the SA receives the learned experience cards.

\begin{center}
\begin{minipage}{\columnwidth}
\centering
{\small
\setlength{\tabcolsep}{3.5pt}
\begin{tabular}{lcccc}
\toprule
Condition & W/L & Win rate & Mean end & Median end \\
\midrule
Before learning & 3/7  & 30\%  & 28,173.1 & 22,800.0 \\
After learning & 10/0 & 100\% & 20,842.8 & 17,380.5 \\
\bottomrule
\end{tabular}
}
\captionof{table}{Full-match results before and after learning under a fixed scenario; end times are in ticks.}
\label{tab:before_after_results}
\end{minipage}
\end{center}

\subsection{Experience Learning under Different Play Styles}

\begin{table*}[!t]
\centering
{\small
\setlength{\tabcolsep}{3pt}
\renewcommand{\arraystretch}{1.08}
\begin{tabularx}{\textwidth}{
    p{0.16\textwidth}
    >{\raggedright\arraybackslash}X
    >{\raggedright\arraybackslash}X}
\toprule
Experience theme & Rush condition & Turtle condition \\
\midrule

Vehicle production
&
Lower the production threshold under a thin economy. With fewer than four harvesters, allow cash to fall to about 100 and queue one vehicle at a time to prevent the war factory from idling.
&
Use stage-dependent cash floors. Keep about 200 before the second refinery. Afterward, produce vehicles while cash exceeds 1,500 and stop near 800.
\\

Economic and production expansion
&
Prioritize the second refinery over additional production capacity, radar, and technology to stabilize the basic economy before further development.
&
Under low pressure, prioritize a third refinery and another war factory so that economy and production capacity expand together.
\\

Post-technology attack
&
Produce advanced units in short batches and wait for an attack window.
&
Reduce prolonged rallying and waiting. When at least six combat units are available and few opponents are visible, the SA is more likely to issue a proactive attack directive.
\\

\bottomrule
\end{tabularx}
}
\caption{Main revisions produced from the same initial experience cards under Rush and Turtle conditions.}
\label{tab:style_cards}
\end{table*}

Rush AI and Turtle AI expose different process-level problems. Against Rush AI, 7 of 10 matches contain long periods with production capacity but insufficient cash, and 6 of 10 show competition between the technology chain and light-unit production. Early pressure therefore forces a limited economy to fund combat, production, and technology at the same time. Against Turtle AI, 7 of 10 matches show vehicle-queue congestion together with idle infantry production, and 7 of 10 retain unspent cash in the middle or late game. Five of 10 miss opportunities for economic expansion under low pressure, and 4 of 10 lack advanced or siege units in the late game. Compared with Rush AI, the Turtle condition makes less effective use of safe windows and converts economic resources into production capacity and attacks less effectively.

These process differences also change the content of formal experience cards. Both conditions end with 16 formal cards, but three cards develop different versions. Table~\ref{tab:style_cards} summarizes the main revisions.

The RA also proposes new candidates. Against Rush AI, one candidate recommends building the second refinery before advancing technology when pressure is low and only one refinery is available. Against Turtle AI, another candidate recommends using an MCV to establish a second base under sustained low pressure while maintaining a minimum basic force before advancing technology. These candidates address development order under a thin economy, expansion during safe windows, and the balance between technology and basic defense.

At the end of each ten-match sequence, each new candidate has only one supporting observation and therefore remains in the candidate pool. The pool retains single-match strategic proposals until cross-game evidence supports promotion.

\begin{center}
\includegraphics[width=0.88\columnwidth]{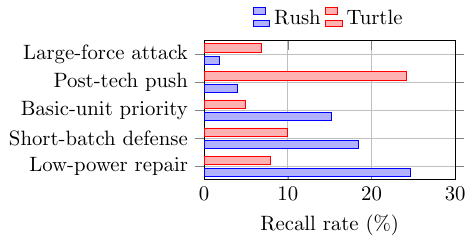}
\captionof{figure}{Percentage of SA decisions in which representative formal experience cards were retrieved under Rush and Turtle conditions. One decision may retrieve multiple cards.}
\label{fig:card-retrieval}
\end{center}

Figure~\ref{fig:card-retrieval} compares the retrieval frequencies of representative formal experience cards. Against Rush AI, the SA more often retrieves cards for low-power recovery, short-batch defense, and basic-unit priority. Against Turtle AI, it more often retrieves cards for post-technology quality advances and large-force attacks.

Overall, the experience learned against Rush AI increasingly favors continued production, defense, and economic recovery under a thin economy. The experience learned against Turtle AI emphasizes expansion under low pressure, resource conversion, post-technology quality advances, and proactive attacks. The process diagnoses, formal experience cards, and SA retrieval distributions point in the same direction. Together, they show that the same initial experience develops into distinct long-term strategic preferences under different opponent styles.

\section{Conclusion and Limitations}

We present STRATA, a role-aligned hierarchical system with cross-game self-learning. The SA makes strategic decisions and issues high-level directives, the LA and TA execute logistics and tactical actions, and the RA derives candidate experience from full game traces and continually validates and revises experience cards. In Red Alert, STRATA forms 16 formal cards over 80 matches against Normal AI. Under the fixed scenario, using these cards increases the observed win rate from 30\% to 100\%. Sequential learning against Rush and Turtle opponents also produces distinct long-term strategic preferences.

The experiments use one map, one fixed seed, one decision model, and a limited number of matches. Future work will expand these settings and evaluate the real-time performance and generalization of alternative role organizations, decision frequencies, and experience-update strategies.

{\small
\bibliography{references}
}

\end{document}